# End to End ASR System with Automatic Punctuation Insertion

CSC2518H Spoken Language Processing

Gavin Guan

# Abstract

Recent Automatic Speech Recognition systems have been moving towards end-to-end systems that can be trained together [1] [2] [3]. Numerous techniques that have been proposed recently enabled this trend, including feature extraction with CNNs, context capturing and acoustic feature modeling with RNNs, automatic alignment of input and output sequences using Connectionist Temporal Classifications, as well as replacing traditional n-gram language models with RNN Language Models [4] [5] [6] [3]. Historically, there has been a lot of interest in automatic punctuation in textual or speech to text context. However, there seems to be little interest in incorporating automatic punctuation into the emerging neural network based end-to-end speech recognition systems, partially due to the lack of English speech corpus with punctuated transcripts. In this study, we propose a method to generate punctuated transcript for the TEDLIUM dataset using transcripts available from ted.com [7] [8]. We also propose an end-to-end ASR system that outputs words and punctuations concurrently from speech signals. Combining Damerau Levenshtein Distance and slot error rate into DLev-SER, we enable measurement of punctuation error rate when the hypothesis text is not perfectly aligned with the reference [9] [10]. Compared with previous methods, our model reduces slot error rate from 0.497 to 0.341 [11].

# 1 Introduction

In recent years, end-to-end Neural Network based ASR systems have become increasingly popular, outperforming the Hidden Markov Models (HMM) and Gaussian Mixture Models (GMM) that were previously state-of-the-art [12]. Inserting punctuation automatically has drawn decent attention prior to the emergence of end-to-end ASR systems [13] [14] [15] [16]. Many of the systems predicted punctuation using combined prosodic features and n-gram language models. The most widely used prosodic features are pause and pitch and they have shown to be relatively effective. However, the use of such prosodic features still imposes an information bottleneck where only the pause and pitch around the potential location for inserting punctuation are considered. More subtle information, such as the overall pace of the sentence, pitch variation of the entire sequence would sometimes be discarded.

There seems to be less interest in automatic punctuation in ASR systems since the emergence of end-to-end Neural Network systems. We think part of the reason might be lack of a large amount of corpus with punctuated transcripts.

In our work, we propose a method to regenerate ground truth transcript with punctuation using the TEDLIUM corpus and transcripts available from ted.com. We also present our model's architecture which involves the use of 2D Convolutional Neural Network (CNN) feature extractors, bidirectional Long Short-Term Memory (LSTM) layers, and Connectionist Temporal Classification (CTC). In addition, we also introduce a new set of CTC Tokens, suitable for predicting punctuations directly from the speech signal. Instead of using n-gram language models that were widely used in other works, we chose to use an LSTM based language model. Lastly, we propose an improved calculation for Slot Error Rate (SER) that enables the calculation of SER of punctuations when the hypothesis transcript is not exactly aligning with the reference.

# 2 Previous Work

Studies on automatic punctuation used a combination of prosodic features and n-gram language models, and sometimes lexicon-based features. The prosodic features are often hand selected and separately calculated, and later used to perform automatic punctuation. Shriberg Et al. used pauses, intonations (F0 features), speaking rate and other features such as word position to construct decision trees [13]. N-gram language models were used to further improve the accuracy [13]. Kolar Et al. similarly combined Prosodic features and n-gram language models, and treated phoneme duration as a feature, which is similar to the speaking rate feature from

Shriberg [14]. Christensen Et al. extracted similar features, and used a finite state machine (FSM) to generate puncutations [15]. He also studied the performance of automatic feature extraction with multilayer perceptron [15]. However, the multilayer perceptron's performance did not exceed that of the FSM's [15]. Batista combined many kinds of prosodic features and lexical features and restored punctuations using Hidden Markov Models (HMMs) and max entropy [16].

Tilk Et al. experimented Punctuation Restoration with LSTMs on speech transcript and pause duration features extracted from the speech signal [11]. However, their system did not attempt to perform speech recognition. In another work from Tilk Et. Al, a bidirectional Gated Recurrent Unit with attention mechanism model is used [17]. However, the system still does not perform speech recognition, automatic punctuation is done on the transcribed text. Similarly, Salloum Et al. performed punctuation restoration on medical transcripts using Bidirectional RNN and Attention layers [18].

# 3 Test Dataset

## 3.1 The TEDLIUM Corpus

We chose the TEDLIUM corpus because it has continuous audio files that contain recordings of multiple sentences (i.e. one entire TED Talk in each file) [7]. Many other popular corpora such as LibriSpeech and WSJ Corpus contain recordings of individual sentences, which makes them inherently unusable for our work [19] [20].

Each recording in TEDLIUM Corpus is structured in:

> *SpeakerNameYear.sph: audio recording of one TED Talk.*
> *SpeakerNameYear.stm: text file containing segmented transcript and their starting and end time.*

The original transcript included in the dataset is punctuation-less. However, the full punctuated transcript for many of the audio files included in the dataset are available from ted.com. Knowing the speaker for each TED talk, web scraping was done to retrieve the transcripts whenever possible. The following steps are performed:

(1) Retrieve all the transcripts for each speaker from ted.com.

(2) Select the most similar punctuated transcript.

Temporarily remove all punctuation marks from the transcripts retrieved online, and calculate a ratio based on Levenshtein Distance between the online transcript and the transcript in the stm file. The ratio is calculated using dividing the Levenshtein distance between the online and stm transcript by the length of the transcript in stm file. The online transcript with the highest ratio is selected as the punctuated transcript if its ratio is greater than 0.6. If no ratio is greater than 0.6, the recording is simply discarded.

(3) Cleanup the punctuated transcript

Remove unnecessary information, or substitute information with their more suitable form. In the following table, we summarize the cleanup operations we have performed.

| Operation | Old Form | New Form |
| --- | --- | --- |
| Remove speaker tags and Information tags | `[Speaker A]`<br>`[Speaker B]`<br>`[Video]`<br>`[Music]`<br>`[Laughter]` | |
| Substitute special characters that are rare with their closest form or pronunciation | `æ`<br>`²`<br>`°`<br>`%`<br>`*` | `ae`<br>`square`<br>`degree`<br>`percent`<br>`times` |
| Convert numbers to their pronunciations | `21st`<br>`1000` | `twenty-first`<br>`one thousand` |
| Remove the dot in acronyms | `st.`<br>`.etc` | `st`<br>`etc` |
| Convert all punctuations to one of comma, period, question mark, exclamation mark and apostrophe. Detach apostrophe from first part of the word. | `: ,`<br>`… .`<br>`!`<br>`?`<br>`didn't` | `,`<br>`.`<br>`!`<br>`?`<br>`didn 't` |
| Separate comma, period, question mark and exclamation mark from the word if it is attached to one. | `Hello. This is` | `Hello . This is` |
| Convert to lower case | `We`<br>`DNA` | `we`<br>`dna` |

Table 1. List of Transcript Cleanup Operations

(4) Perform Pairwise alignment between the punctuated transcript and the unpunctuated transcript.

Introduced by Stephen F Altschul and Warren Gish, the basic local alignment search tool was originally a tool designed for aligning similar DNA and protein sequences [21]. We used its Python open source information available from the module pairwise2 [22]. Different scores can be assigned for "identical characters", "substitution for non-identical characters", "opening a gap for either the reference or hypothesis (i.e. introducing insertion or deletion error)", or "extending a gap", we assign 1, -1, -5 and -0.01 for each of them for our alignment task. We introduce a high penalty for "opening a gap" because the stm transcript in the original corpus can have mistakes and often do not contain the entire transcript. Therefore, the program tends to introduce a lot of gaps if to align identical characters if the cost for opening a gap is low. Consider the following example where the stm transcript contains an extra "the" that was not expected.

```
this is the development plan for asia. we believe
th--------e-------------------------we believe
```

Figure 1. Example Effect of Spurious Word with Low Open Gap Penalty

By introducing a high penalty for opening a gap, we force the erroneous word to stay close to the rest of the sequence, making the alignment easier.

(5) Replace the transcript in the stm files with their punctuated version.

The final stm has the format similar to:

```
[1.56] [10.56] this is a nine second segment. Each sentence
[10.86] [15.73] may be segmented into multiple audio pieces
```

Figure 2. Example stm File with Punctuated Transcript

### 3.2 The Wikipedia Corpus

We used the Wikipedia text corpus to train our Long Short-Term Memory Language Model, because most of the ready to use language models are being trained without punctuation [23]. We selected pieces from the Wikipedia corpus based on the following heuristics:

(1) Only the main bodies of text (paragraphs) in Wikipedia corpus are kept. Titles, section titles and captions are eliminated.
(2) Cleanup similar to the TEDLIUM transcript is applied whenever applicable.
(3) Eliminated paragraphs that are overly short: containing less than two full stops (. , !) or 20 words.
(4) Eliminated sentences that contained non-alphanumerical symbols (characters not in [a-zA-Z0-9] or <unk>).
(5) Sections that contained no more than two consecutive sentences as a result of step (4) are also eliminated.

We kept the <unk> in the text body.

# 4 Model Architecture

In this section, we provide an overview of our model's architecture. Our final model has a audio preprocessing component, 2D CNN layers, BiLSTM layers, a fully connected layer and a CTC layer.

## 4.1 Audio Signal Preprocessing

In recent years, the use of Fourier Transform (FT) instead of applying the full Mel-frequency Cepstral Coefficients (MFCC) become more popular [3] [24]. This is partly due to the enhanced capability of end-to-end models so that there is less a need for a fine preprocessed audio signal. In our work, we applied Short Time Fourier Transform (STFT) on the audio signal with a window of 20ms and stride of 10ms. No further processing such as taking cepstrum and adding delta features are taken [25]. The logarithm of the resulting spectrum is ultimately fed into the model.

## 4.2 2D Convolutional Neural Network (2D CNN)

Using Convolution Neural Network as feature extractors has become a common practice for end-to-end Automatic Speech Recognition systems. Multiple works have reported an improvement in WER when CNN feature extractors are used [24] [26].

We adapted the 2-layer 2D CNN from Baidu's Deepspeech 2, the final output of the CNN layers has 32 channels. Its detailed parameters are listed at the end of this section. We didn't further experiment with different configurations because its exact configuration seems to have a very minor effect on the accuracy according to the Deepspeech 2 paper [24].

## 4.3 Bidirectional Long Short Term Memory (BiLSTM)

At the core of our system are multiple bidirectional Long Short Term Memory (BiLSTM) layers stacked together. Given an input sequence $x = (x_1, \ldots x_T)$, a typical Recurrent Neural Network (RNN) would compute a hidden vector sequence $h = (h_1, \ldots, h_T)$ using the input sequence, and compute the output sequence $y = (y_1, \ldots y_T)$ using the hidden vector sequence [1].

$$h_t = H(W_{ih}x_t + W_{hh}h_{t-1} + b_h)$$

$$y_t = W_{ho}h_t + b_o$$

In our model, the input sequence to the recurrent layers has 32 channels, as a result of feature extraction from the 2D CNN layers.

Compared to a vanilla RNN network, in an LSTM network, the hidden layer activation function H, is replaced by [27]:

$$Input\ Gate: i_t = \sigma(W_{xi}x_t + W_{hi}h_{t-1} + W_{ci}c_{t-1} + b_i)$$

$$Forget\ Gate: f_t = \sigma(W_{xf}x_t + W_{hf}h_{t-1} + W_{cf}c_{t-1} + b_f)$$

$$Output\ Gate: o_t = \sigma(W_{xo}x_t + W_{ho}h_{t-1} + W_{co}c_t + b_o)$$

$$Activation: c_t = f_t c_{t-1} + i_t \tanh(W_{xc}x_t + W_{hc}h_{t-1} + b_c)$$

$$Hidden\ Layer\ Output: h_t = o_t \tanh(c_t)$$

Compared to an LSTM network, a BiLSTM network involves a forward hidden sequence $\overrightarrow{h_t}$, and a backward hidden sequence $\overleftarrow{h_t}$ [28]. A BiLSTM is defined as [1]:

$$\overrightarrow{h_t} = H(W_{x\overrightarrow{h}}x_t + W_{\overrightarrow{h}\overrightarrow{h}}\overrightarrow{h}_{t-1} + b_{\overrightarrow{h}})$$

$$\overleftarrow{h_t} = H(W_{x\overleftarrow{h}}x_t + W_{\overleftarrow{h}\overleftarrow{h}}\overleftarrow{h}_{t-1} + b_{\overleftarrow{h}})$$

$$y_t = W_{\overrightarrow{h}y}\overrightarrow{h_t} + W_{\overleftarrow{h}y}\overleftarrow{h_t} + b_o$$

where H is the LSTM hidden layer activation function.

We chose an LSTM network due to its capability in finding long term dependency patterns. It can selectively remember and forget information. We found that the detection of question marks relies heavily on this capability of LSTM. While the detection of terminal punctuation (. ! ?) may rely on pauses at the location of the punctuation, for which RNNs have also been doing a decent job. Correctly classifying question marks from periods relies heavily on remembering pitch and tone from earlier sections of a sentence. In an RNN model, most question marks are mistakenly classified as periods. All other metrics such as Word-Error-Rate (WER) and other punctuation error rate all benefited from the adaptation of LSTM network as well, to less of an extent.

Bidirectional LSTM is chosen in favor of LSTM because it can use context from both sides to detect punctuation. The clue for punctuation often appears after the punctuation itself. For example, terminal punctuations often exist before words such as "I, We, Where".

## 4.4 Connectionist Temporal Classification (CTC) and LSTM Language Model

The original CTC token set contains the English alphabet, as well as space, apostrophe and a blank token [6]. The entire set has 29 tokens in total. Our model instead uses 33 tokens, namely:

`Original CTC Tokens + {COMMA, PERIOD, QUESTION_MARK, EXCLAMATION_MARK}`

For training, we chose an off shelf parallel implementation of CTC loss that can be run on GPUs [24]. During inference, we combined a separately trained LSTM Language Model (LSTM-LM) with beam search to improve the model's accuracy [6] [29]. It treats punctuation marks as individual work tokens. The LSTM-LM has two layers of LSTMs with a hidden size of 200.

Our final decoding objective is:

$$c = \underset{c}{\mathrm{argmax}}\{\log(p_{ctc}(c|x)) + \gamma \log(p_{lm}(c))\}$$

where $p_{lm}(c)$ is the probability of the word sequence c given the LSTM Language Model

$$p_{lm}(c) = \prod_{i=1}^{L} p_{lm}(c_i|c_1, \dots, c_{i-1})$$

## 4.5 Training Scheme

We trained our model with Nesterov's Accelerated Stochastic Gradient Descent with a momentum of 0.9 and weight decay of 1e-5 [30]. We chose a batch size of 10 due to GPU memory constraints. There wes a total of 70 epochs in training while the loss stopped decreasing at around 35$^{th}$ epoch. The following diagram shows the loss of the model during training.

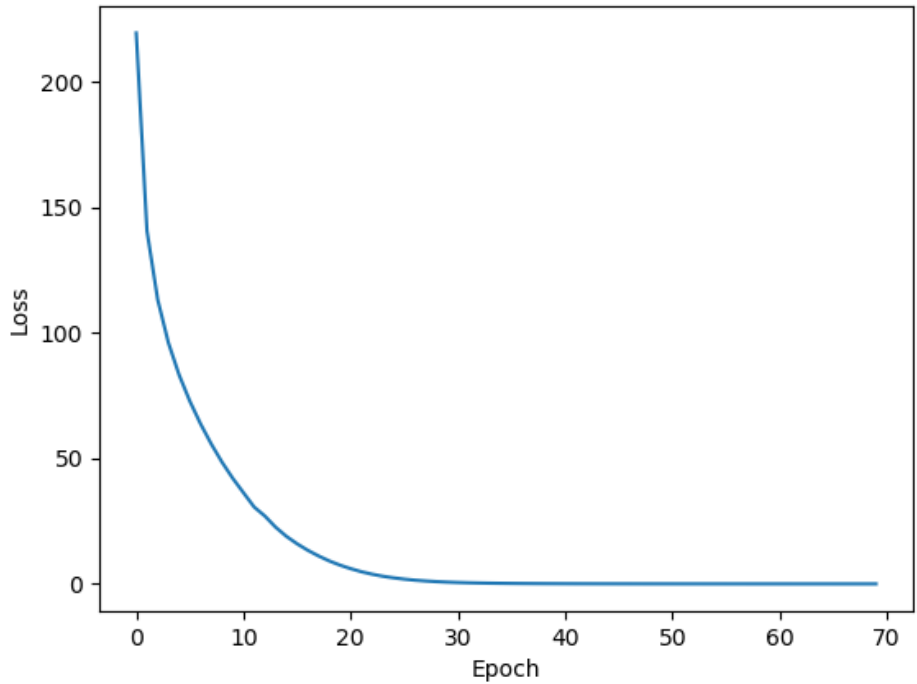

Figure 3. Loss of the Model vs. Training Epoch

## 4.6 Architecture Summarization

The final architecture is summarized in the following table.

| Layer Type | Characteristics |
| --- | --- |
| 2D Convolutional Neural Network | 1st layer: 32 output channel; (41, 11) kernel size; (2, 2) stride; (20, 5) padding <br> 2nd layer: 32 output channel; (21, 11) kernel size; (2, 1) stride; (10, 5) padding |
| Bidirectional Long-Short Term Memory | 5 identical layers with hidden size of 768. |
| Fully Connected Layer | Input feature size: 768. Output feature size: 33 |
| Connectionist Temporal Classification (With LSTM-LM for inference) | LSTM-LM has 2 identical layers with hidden size 200. |

Table 2. Summarization of Model Architecture

# 5 Punctuation Error Rate Metrics

There has yet been a common metrics for evaluating punctuation error rates. One method is to evaluate using Precision, Recall, and the F-score [31]. Another popular method is the Slot Error Rate (SER), which calculates the ratio between the sum of insertion, deletion and substitution error and the number of instances in the reference [9]. We will briefly present these two methods in section 5.1 and 5.2. To the best of our knowledge, none of the previous automatic punctuation works clearly explained how the F-measure or SER for the punctuations was calculated given the word tokens in the predicted text may not exactly align with those in the reference text. In section 5.3, we will propose a new metric DLev-SER which combines Damerau-Levenshtein Distance and Slot Error Rate (DLev-SER) [9] [10]. DLev-SER can calculate punctuation error rate given misalignments in the words and can propose a punctuation error rate that is independent of the correctness of the alphanumerical text of the hypothesis. In section 5.4, we propose a method to calculate DLev-SER for different types of punctuations individually to better evaluate the system's performance on each one of them.

## 5.1 Precision, Recall and F-score

$$Precision = \frac{Number\ of\ Correct\ Punctuations}{Number\ of\ Punctuations\ in\ Hypothesis}$$

$$Recall = \frac{Number\ of\ Correct\ Punctuations}{Number\ of\ Punctuations\ in\ Reference}$$

$$F = \frac{2PR}{P + R}$$

Precision represents the percentage of correct instances among all instances in the hypothesis, and recall is the correct instances that have been retrieved for all instances in the ground truth (reference). The F-score seeks to balance Precision and Recall.

## 5.2 Slot Error Rate (SER)

Precision, Recall and F-score have been popular evaluation metrics for many automatic punctuation solutions [9]. However, it has also been criticized for deweighting missing and extra instances by a factor of two compared to substitutions. Slot Error Rate seeks to equally value insertion, deletion and substitution errors. SER is commonly obtained by:

$$SER = \frac{I + D + S}{C + S + D} = \frac{All\ Punctuation\ Errors}{Number\ of\ Punctuations\ in\ Reference}$$

Where C is the correctly identified punctuation, and I is the insertion error, D is the deletion error. S is the substitution error - a different type of punctuation mark other than the correct one is predicted.

The Slot Error Rate for punctuations is easy to calculate if the words are perfectly aligned between the reference and the hypothesis. However, to the best of our knowledge, no previous work has discussed how to handle the following common situation:

```
Ref   w1    w2   ,    w4

Hyp   w1    .    w4
```

Figure 4. Example Reference and Hypothesis Word Mismatch

Whether the period in the hypothesis should be considered as a substitution error (1 error), or should be considered as an insertion error while the missing comma is another error (2 errors) is subject to discussion. Thus, we propose a new and rigorous method for calculating the SER rate in the next section.

### 5.3 Damerau-Levenshtein Slot Error Rate

Both the PR&F metrics and generic SER work assuming perfect word alignment between the - reference and hypothesis. However, the exact method to calculate these numbers become unclear if the words are not aligned. Levenshtein distance is a method that calculates the minimal edit distance between two sequences, each edit may be an insertion, deletion or substitution of an element. Damerau-Levenshtein distance differs from the previous method by treating the swap of two elements as one edit, which would take require two edits otherwise [10].

We propose a new Damerau-Levenshtein Slot Error Rate (DLev-SER). For DLev-SER, we put elements in transcript into two categories: words and punctuations. We first calculate the minimal edit distance between the reference and the hypothesis, subject to rules below. The Slot Error Rate is then calculated based on how many errors occurred on the punctuations.

In detail, we use the following rules when calculating DLev-SER:

(1) A word cannot be substituted with a punctuation, or vice versa.
(2) A punctuation can swap position with a word, which is considered a swap (W) for punctuation.

(3) In addition, a punctuation can be inserted (I), deleted (D), substituted by another punctuation (S) or swap position with another punctuation (W).

(4) A word can be inserted, deleted, substituted by another word, or swap position with another word as usual, contributing to the minimal Damerau-Levenshtein distance calculation but is not included in DLev-SER calculation.

(5) A punctuation insertion, deletion, substitution or swap is considered an edit with cost 0.999. An edit on a word has cost 1.0.

The final punctuation error rate is given by:

$$DLevSER = \frac{All\ Punctuation\ Errors}{Number\ of\ Punctuations\ in\ Reference} = \frac{I + D + S + W}{C + S + D}$$

We selected the Damerau-Levenshtein distance over the Levenshtein distance because we found sometimes the hypothesis predicted punctuations at locations very close (off by one word) to the ground truth. Without the swap edit, such prediction would require an insertion and a deletion. We feel penalizing such a close prediction twice over other more severe mistakes such as introducing a spurious punctuation is unfair.

There are two edge cases we have considered:

1. Spurious punctuation introduced when there was no punctuation in the reference.

Many of the previous works did not mention what the SER would be in this case. In our calculation, we chose to make the "Number of Punctuations in Reference" to be 1. Therefore, if there are 2 insertion errors in the hypothesis while no punctuation is present in the reference, the DLev-SER is 2. If there are punctuations in the reference, we do not add a 1 to the denominator. This is due to the additional 1 in the denominator will reduce the DLev-SER, giving us an unfair advantage when comparing with results from previous work.

2. Multiple paths with same number of edits are possible for converting reference into hypothesis.

Consider the following simple case:

```
Ref   w1    .
Hyp   .     w2
```

Figure 5. Example Reference and Hypothesis Word Mismatch

To convert hypothesis to reference, there are two paths with that would require the same minimal number of edits. The first path is to swap PERIOD and w2, and substitute w2 with w1 (1 edit on punctuation + 1 edit on word). The second path is to insert w1 before PERIOD, and delete w2 (0 edit on punctuation + 2 edits on word). This is not an issue for the original definition of Damerau-Levenshtein Distance, since both paths would give the same distance. However, the choice of which path will affect out DLev-SER calculation. Therefore, we introduce rule (5) from above, giving punctuation edits a lower cost (0.999 instead of 1), so that the path with a greater number of punctuation edits will always be selected. Unless there are 1000 punctuation edits possible, this final edit path is still one of the paths with the least number of edits.

By selecting the paths with more punctuation edits, we make our DLev-SER higher. This is also to make sure we don't have unfair advantage when results are compared with previous work.

### 5.4 Punctuation Specific Damerau-Levenshtein Slot Error Rates

We further define DLev-SER rate for comma, period and question marks individually. They are defined similarly as DLev-SER for all punctuations. The only difference is that only one punctuation is treated separately from the rest of the words and punctuations. Take question mark DLev-SER for example, inserting, deletion, substituting a question mark with other punctuation or words, as well as swapping a question mark with other tokens contribute to question mark DLev-SER. Substituting a comma with a period would not contribute to question mark DLev-SER.

We do not calculate exclamation mark DLev-SER separately because we found our model almost never predicted exclamation mark, they are typically replaced by period or comma in the hypothesis. We believe this is due to the total number of exclamation marks is too small, and the prosodic features are not strong enough to overcome this lack of training samples.

## 6 Result and Discussion

### 6.1 Comparison of Effectiveness of Different Layers

In this section, the performance of our model compared with its own variants, as well as with previous works will be presented. We will compare both Word Error Rate (WER) and Character Error Rate (CER). We will also compare our model's DLev-SER with previous work's SER, and compare our individual punctuation DLev-SER whenever possible. Unfortunately, we could not find a way to accurately define punctuation precision, recall and f-score for our task so that we will not compare this metric against previous work.

|  | WER | CER | DLev-SER | ? DLev-SER | . DLev-SER | , DLev-SER |
|---|---|---|---|---|---|---|
| RNN Model | 35.526 | 17.153 | 0.721 | 0.489 | 0.418 | 0.896 |
| BiLSTM Model | 26.263 | 8.160 | 0.567 | 0.247 | 0.270 | 0.694 |
| BiLSTM Model + LSTM LM | 10.877 | 5.618 | 0.341 | 0.259 | 0.251 | 0.446 |

Table 3. Comparison of Performance for Different Feature Combinations

As shown in Table, we found our BiLSTM variant is outperforming the RNN variant in every aspect. It is worth noting that question mark prediction seems to benefit the most from adapting BiLSTM layers. With BiLSTM layers, the error rate for punctuation detection is actually lower than both period DLev-SER and comma DLev-SER. This is due to the prosodic features for detecting question mark is typically strong and have less ambiguity. However, the features are less localized and may be hard for the RNN cells to keep track of.

Adding an LSTM-LM further improves performance in most categories. Comma DLev-SER benefited the most from the introduction of a LSTM-LM. This is most likely due to commas can occur close to conjunction words, which presents both in TED talks and Wikipedia corpus. In addition, many spurious commas are inserted when short pauses in the talk are detected. With the assist of a language model, many obviously spurious are eliminated.

The LSTM-LM improved period DLev-SER to a lesser extent. The period DLev-SER is already lower compared to that of comma for the model without LM, this is because the prosodic clue for a period is typically stronger than a comma. Periods are often being predicted at places with long pauses and followed by other sentence starting keywords. The LSTM-LM could not drastically improve the period DLev-SER because the content in Wikipedia and TED is often very different. Wikipedia text is more objective than TED talks, sentences often start with proper nouns. TED talks instead have many sentences starting with "We, They, Where, If", which are rare for the Wikipedia text. We believe if the LSTM-LM can be trained with text that is more similar to the TED talk itself, the error rate can be further reduced.

For a similar reason, the LSTM-LM did not improve question mark detection. It actually made some correctly detected punctuation mark into period or comma. Due to the nature of Wikipedia text, question marks hardly exist.

## 6.2 Comparison with Previous Studies

In the following table, we compare our model's performance with some previous works. Note that earlier work's calculated generic SER instead of DLev-SER since they are working on a perfect transcript.

|  | DLev-SER | ? DLev-SER | . DLev-SER | , DLev-SER |
|---|---|---|---|---|
| Tilk's BiLSTM + Attention [11] | 0.497 | NA | NA | NA |
| Batista's HMM [16] | 0.695 | NA | 0.361 | 1.07 |
| BiLSTM Model + LSTM LM (Our Model) | 0.341 | 0.259 | 0.251 | 0.446 |

Table 4. Comparison of Performance with Previous Studies

When comparing our result with the results of previous works that performed punctuation restoration using selected prosodic features, we found our model is performing better. Previous studies often worked with transcribed text, and uses a few prosodic features extracted from the audio signal. Although this means they can lose certain useful feature from the speech, but also gives them the advantage of working with a correct transcript. Our system instead produces the transcript and punctuations concurrently, and we specifically designed the DLev-SER to treat as many errors as punctuation errors as possible. With all factors considered, our system proved to be a big step in terms of reducing error rate for automatic punctuation.

# 7 Conclusion

In our work, we trained and tested a BiLSTM based end-to-end automatic speech recognition system with automatic punctuation capability. The system uses many state-of-the-art features including 2D CNN feature extractor, multilayer BiLSTM, CTC and LSTM-LM. We proposed a set of CTC tokens that are suitable for generating punctuation marks directly from the speech signal. We also developed a new method for calculating the slot error rate for punctuations with potentially misaligned text using Damerau-Levenshtein Distance. We see the improved result in slot error rate when compared to previous studies that use specific prosodic features. The overall slot error rate from all punctuations combined reduced from 0.497 to 0.341.